\documentclass{article}

\usepackage[preprint,nonatbib]{neurips_2024}

\usepackage[pagebackref,breaklinks,colorlinks]{hyperref}
\usepackage[utf8]{inputenc} 
\usepackage[T1]{fontenc}    
\usepackage{hyperref}       
\usepackage{url}            
\usepackage{booktabs}       
\usepackage{amsfonts}       
\usepackage{nicefrac}       
\usepackage{microtype}      
\usepackage{xcolor}         
\usepackage{multirow}
\usepackage[pdftex]{graphicx}
\usepackage{amsmath}
\usepackage{algorithm}
\usepackage{algorithmic}
\usepackage{bbding}
\usepackage{pifont}
\usepackage{makecell}
\usepackage{subfig}

\usepackage{latexsym}
\usepackage{amssymb}

\definecolor{codegreen}{rgb}{0,0.4,0}
\definecolor{codegray}{rgb}{1.0,0.5,0.5}
\definecolor{codepurple}{rgb}{0.58,0,0}
\definecolor{tealblue}{rgb}{0,0.5,0.5}
\definecolor{codebackcolour}{rgb}{0.95,0.95,0.92}
\definecolor{darkgreen}{RGB}{0,127,0}
\definecolor{darkred}{RGB}{200,0,0}
\definecolor{orange}{rgb}{1,0.5,0}
\def\greencheckmark{\textcolor{darkgreen}{\checkmark}}
\def\redxmark{\textcolor{darkred}{\ding{55}}}  

\definecolor{grey50}{rgb}{0.5,0.5,0.5}

\title{Storynizor: Consistent Story Generation via Inter-Frame Synchronized and Shuffled ID Injection}

\author{
Yuhang Ma\textsuperscript{1}$^\ast$\thanks{Equal Contribution}\thanks{Project Lead},
~~~ Wenting Xu\textsuperscript{1}$^\ast$,
~~~ Chaoyi Zhao\textsuperscript{1}$^\ast$,
~~~ Keqiang Sun\textsuperscript{2},\\
~~~ \bf{Qinfeng Jin}\textsuperscript{1},
~~~ \bf{Zeng Zhao}\textsuperscript{1}\thanks{Corresponding Authors},
~~~ \bf{Changjie Fan}\textsuperscript{1},
~~~ \bf{Zhipeng Hu}\textsuperscript{1}
\\
\textsuperscript{1}Fuxi AI Lab, NetEase Inc.\\ 
\textsuperscript{2}Multimedia Laboratory, The Chinese University of Hong Kong\\
}

\begin{document}

\maketitle
\begin{abstract}
\noindent
Recent advances in text-to-image diffusion models have spurred significant interest in continuous story image generation. In this paper, we introduce \textbf{Storynizor}, a model capable of generating coherent stories with strong inter-frame character consistency, effective foreground-background separation, and diverse pose variation. The core innovation of Storynizor lies in its key modules: \textbf{ID-Synchronizer} and \textbf{ID-Injector}. The ID-Synchronizer employs an \textit{auto-mask self-attention} module and a \textit{mask perceptual loss} across inter-frame images to improve the consistency of character generation, vividly representing their postures and backgrounds. The ID-Injector utilize a \textit{Shuffling Reference Strategy (SRS)} to integrate ID features into specific locations, enhancing ID-based consistent character generation. Additionally, to facilitate the training of Storynizor, we have curated a novel dataset called \textbf{StoryDB} comprising $100,000$ images. This dataset contains single and multiple-character sets in diverse environments, layouts, and gestures with detailed descriptions. Experimental results indicate that Storynizor demonstrates superior coherent story generation with high-fidelity character consistency, flexible postures, and vivid backgrounds compared to other character-specific methods. 
\end{abstract}    
\section{Introduction}

Recent advancements in text-to-image diffusion models has sparked considerable interest in generating continuous story images. Maintaining consistency between frames, ensuring natural and flexible character poses, and achieving a clear separation of foreground and background are critical challenges in this domain.

Many prior works have paid attention to ensuring character consistency. For instance, IP-Adapter~\cite{ipadapter}, Arc2Face~\cite{papantoniou2024arc2face}, and InstantID~\cite{wang2024instantid} extract identity features from a reference image and inject them into the diffusion model. While effective in single-character scenarios, these methods often struggle with stiff postures and are limited in handling more complex multi-character interactions.
\begin{figure}[!t]
    \centering
    \includegraphics[width=0.8\linewidth]{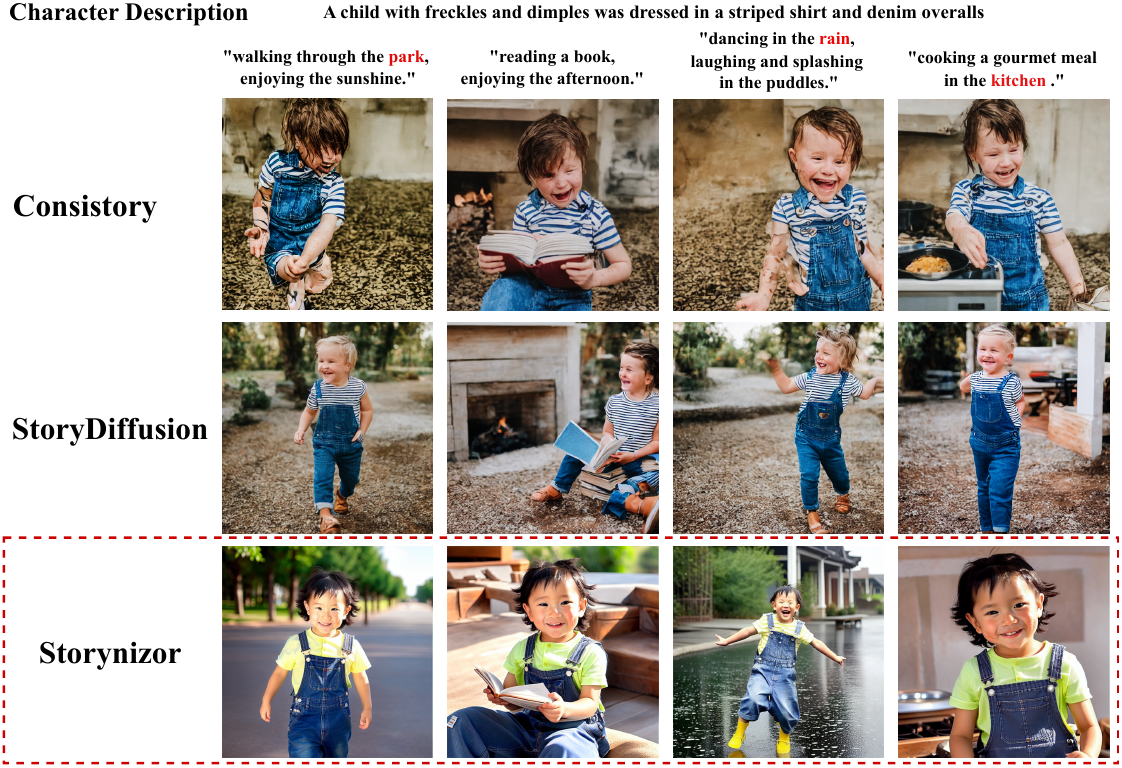}
    \caption{Comparison of Storynizor with existing methods. Storynizor shows superior performance when implemented in the original SD-base checkpoint in text-image alignment and inter-frame consistency.
    }
    %
    \label{fig:introcompare}
\end{figure}

Other approaches, such as Mix-of-Show~\cite{gu2024mix} and OMG~\cite{kong2024omg}, focus on multi-character generation by utilizing attention maps to position characters within a frame. These methods successfully achieve varied poses and maintain character consistency but lack inter-frame coherence, as they operate on a frame-by-frame basis without ensuring consistency across the sequence.

To achieve narrative coherence, methods like ConsiStory~\cite{consistory} and StoryDiffusion~\cite{zhou2024storydiffusion} have attempted to fuse character features across frames to enhance inter-frame consistency. However, the absence of an identity injection mechanism in these approaches results in inaccurate alignment with reference images. Moreover, when considering a pre-trained diffusion model like the original checkpoint of SD1.5, their training-free nature often leads to semantic degradation, and collapsible cross-frame results, as illustrated in Fig. \ref{fig:introcompare}.
\begin{table*}
    \centering
    \small
    \setlength\tabcolsep{0.6mm}{
    \scalebox{0.9}{
    \begin{tabular}{lccccc}
        \toprule
      & ID Consistency  & Flexible Human Pose & Multi-Subject & Inter-Frame Consistency & F/B Disentanglement \\    
         \midrule
         IP-Adapte & \greencheckmark & \redxmark & \redxmark & \redxmark & \redxmark \\
         InctantID & \greencheckmark & \redxmark & \redxmark & \redxmark & \redxmark \\
         OMG & \greencheckmark & \greencheckmark & \greencheckmark & \redxmark & \redxmark \\
         ConsiStory & \redxmark & \greencheckmark & \greencheckmark & \greencheckmark & \redxmark \\
         StoryDiffusion & \redxmark & \greencheckmark & \greencheckmark & \greencheckmark & \redxmark \\
         FastComposer & \redxmark & \greencheckmark & \greencheckmark & \redxmark & \greencheckmark \\
         \textbf{Storynizor (ours)} & \greencheckmark & \greencheckmark & \greencheckmark & \greencheckmark & \greencheckmark \\
         \bottomrule
    \end{tabular}}}
    \caption{Comparison between our proposed Storynizor and state-of-the-art character-specific methods.
    }
        \vspace{-3mm}
    \label{tab:survey}
\end{table*}

As shown in Tab. \ref{tab:survey}, prior works have focused on specific aspects of generating continuous story images, but none of them have comprehensively addressed all the key challenges.

In this paper, we introduce Storynizor, the first model capable of generating multi-character stories with high inter-frame character consistency, effective foreground-background separation, and rich pose variation.

As shown in Fig. \ref{fig:framework}, given arbitrary numbers of reference images and several text prompts from a story, our Storynizor generate corresponding story images, with consistent character identity, vivid character postures and maintaining high consistency across frames.

The core innovation of Storynizor lies in key modules: the ID-Synchronizer to ensure the identity features are consistently maintained across frames, and the ID-Injector to introduce ID-specific features from the reference images.

Specifically, our approach builds upon the UNet architecture, where the ID-Synchronizer, composed of Auto-mask Space-Attention (AMSA) trained with the Mask Perceptual Loss, plays a crucial role in preventing the attention mask leakage and enhancing the consistency of characters throughout the sequence of frames.

In parallel, the ID-Injector extracts essential features from reference characters and integrates them into specific locations within the network.
To make sure the ID-Injector learns the identity information from the reference character images without simply replicating the image feature from the reference image, we introduce a Shuffling Reference Strategy (SRS). Concretely, we randomly sample pairs of reference and ground-truth images from the same character set, with variations in layout, scenarios, and gestures. This strategy significantly boosts the generalization of the model and maintain consistency across diverse poses and environments, leading to notable improvements in performance.

To train Storynizor effectively and support the Shuffling Reference Strategy (SRS), we futher curated a novel dataset, called StoryDB, by selecting multiple sets of characters and collecting images of each character set in various environments, layouts, and gestures. This diverse and carefully structured dataset allows the model to maintain identity consistency while performing different actions in diverse scenarios.


In summary, the contributions of this paper are four folded:
\begin{itemize}

\item We \textbf{introduce Storynizor}, the first model capable of generating multi-character stories with high inter-frame character consistency, effective foreground-background separation, and rich pose variation.
\item We develop two key modules—\textbf{ID-Injector and ID-Synchronizer}—integrated into a UNet-based architecture, ensuring consistent character identity and posture across sequential frames.
\item We curate \textbf{a novel dataset} featuring multiple character sets in various environments, layouts, and gestures, enabling the model to maintain identity consistency across different scenarios and actions.

\end{itemize}

\section{Related work}

\textbf{Text-to-image generative models.} Diffusion models have achieved remarkable results in text-to-image generation in recent years \cite{nichol2021glide,saharia2022photorealistic,ramesh2022hierarchical,ho2020denoising,song2020denoising,balaji2022ediff,ramesh2021zero}. Early works such as DALL-E2 \cite{ramesh2022hierarchical} and Imagen \cite{saharia2022photorealistic} utilize original images as the diffusion input, resulting in enormous computational resources and training time. Latent diffusion models (\textbf{LDMs}) ~\cite{Rombach_2022_CVPR} have been introduced to compress images into a latent space through a pre-trained auto-encoder \cite{van2017neural}, instead of operating directly in the pixel space \cite{saharia2022photorealistic,nichol2021glide}. However, general diffusion models rely solely on text prompts, lacking the capability to generate consistent characters with image conditions.  

\textbf{Consistent character generation.} Subject-driven image generation aims to generate customized images of a particular subject based on different text prompts. Most existing works adopt extensive fine-tuning for each subject \cite{ruiz2023dreambooth,hua2023dreamtuner,hu2021lora,wei2023elite}. Dreambooth \cite{ruiz2023dreambooth} maps the subject to a unique identifier while Textual-Inversion \cite{gal2022textual} is proposed to optimize a word vector for a custom concept. 
Moreover, some works \cite{kumari2023multi,kong2024omg,gu2024mix} put their effort in multi-subject image generation. 
Custom Diffusion \cite{kumari2023multi} propose to combine multiple concepts via closed-form constrained optimization. OMG \cite{kong2024omg} and Mix-of-Show \cite{gu2024mix} propose to optimize the fusion mode during training in circumstance of multi-concept generation. However, these methods necessitate additional training for all subjects, which can be time-consuming in multi-subject generation scenarios. Recently, some methods strive to enable subject-driven image generation without additional training \cite{xiao2023fastcomposer,zhang2023ssr,ipadapter,li2023photomaker,wang2024instantid,consistory}. Most of them explore extended-attention mechanisms for maintaining identity consistency. IP-Adapter \cite{ipadapter} and InstantID \cite{wang2024instantid} introduce visual control by separating cross-attention layers for text features and image features. ConsiStory \cite{consistory} enables training-free subject-level consistency across novel images via cross-frame attention. However, they fail to preserve detailed information according to the inadequate image feature extraction.
\section{Method}
\begin{figure*}[!t]
    \centering
    \includegraphics[width=1\linewidth]{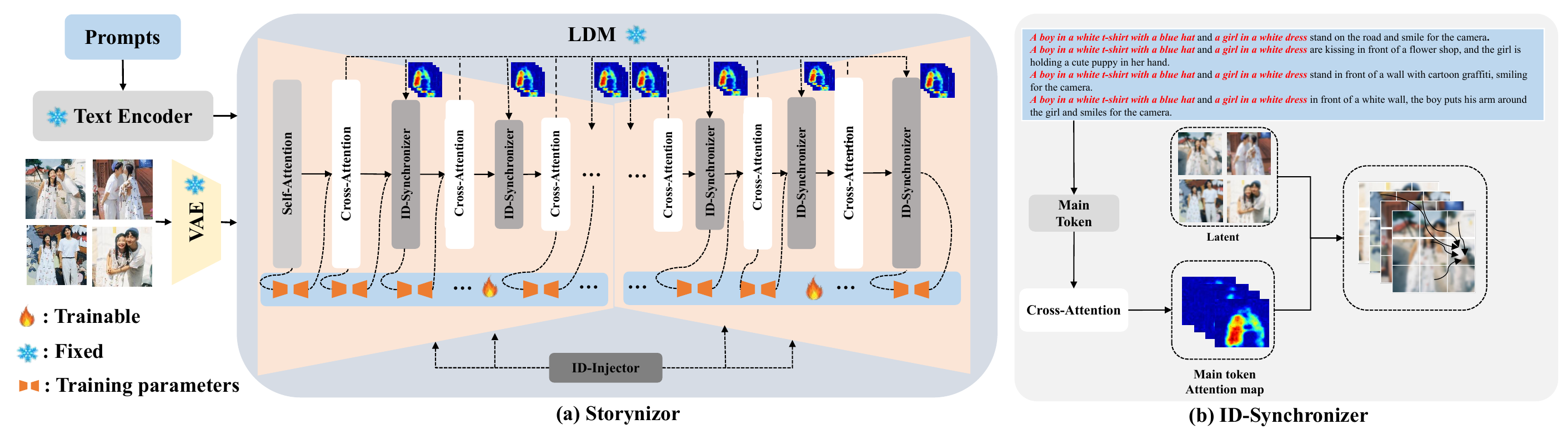}
    \caption{Overview of our proposed (a) Storynizor. Storynizor mainly contains two modules, ID-Injector and ID-Synchronizer. ID-Injector extracts ID features of reference characters with a Shuffling Reference Strategy (SRS), while ID-Synchronizer introduces a mask perceptual loss to modify cross-attention masks and utilizes an auto-mask self-attention module to ensure consistent generation of main characters across inter-frames, as well as vivid background.}
    \label{fig:framework}
\end{figure*}

We propose a pretraining story generation model called Storynizor, which generates a series of multi-character stories with high inter-frame character consistency, effective foreground background separation, and rich pose variation under a series of prompt conditioning and ID images (optional). To modeling our task, we set a series of prompts $\mathcal{T}$ as the following:
\begin{equation}
    \mathcal{T}=\{\mathcal{T}_{n}\}, n = 1,...,N
  \label{eq:textprompt}
\end{equation}
where $N$ denotes the total numbers of prompts. $\mathcal{T}_{n}$ contains the description of characters $P$ and the actions of each character$A$:
\begin{equation}
    \mathcal{T}_{n} = \{P_{n},A_{n}\}=\{P_{n}^{m},A_{n}^{m}\},m=1,...,M,
  \label{eq:Tn}
\end{equation}
where $M$ represents the total number of characters,  $A_{n}^{m}$ represents the action of the $m$-th character in the $n$-th prompt. Notably, $P_{n}^{m}$ refers to the description of the $m$-th character in the $n$-th prompt. Then, the series of multi-character stories generation can be formulated as follows:
\begin{equation}
\begin{split}
    \mathcal{I}_{1},\mathcal{I}_{2},...,\mathcal{I}_{N} = \mathcal{F}(z_{1},...,z_{N}|\mathcal{T},\mathcal{I}_{R},\theta),
\end{split}
\end{equation}
where $z$ denotes the latent noise, $\mathcal{I}_{R}$ represents reference images of characters. $\theta_{i}$ defines the parameters of Storynizor. 

The pipeline of Storynizor is shown in Fig. \ref{fig:framework}(a). In contrast to existing methods, our work makes improvements in two aspects: (1) It consists of an ID-Synchronizer $\mathcal{S}$ which uses an auto-mask spacial attention module to obtain masks during diffusion process, and pay more attention to the character regions across frames, resulting in more precise consistent character and diverse background generation. (2) An ID-Injector $\Phi$ is introduced as a component in Storynizor, which extracts ID features of reference characters and inject it into ID-Synchronizer to achieve image generation with instant Face-ID. 

\subsection{ID-Synchronizer} 
Previous works~\cite{consistory} typically consider a spacial self-attention module to ensure consistency among inter-frames. Given a series of latent noise features $x_{t}\in \mathbb{R}^{B \times F\times H \times W \times C}$ and a single text prompts $y$, they formulate latent noise features as $z_{t}\in \mathbb{R}^{B \times FHW \times C}$ for spacial self-attention to inherit all the module weights from the original 2D self-attention in diffusion model. ID-Synchronizer also begins with this well-explored design. However, the shared visual features across images produce nearly identical backgrounds. While maintaining minimal variation in backgrounds or layout among frames is typical for tasks like video and 3D-object generation, generating narrative images for stories demands vibrant backgrounds tailored to specific text prompts.

Therefore, we introduce an Auto-mask Self-attention (AMSA) to our ID-Synchronizer to ensure consistent character generation in vivid backgrounds and postures. AMSA leverages attention masks of the primary subjects, acquired from the cross-attention modules of the UNet, to concentrate on regions containing characters. It then employs spatial self-attention to these specific areas within the noise features across frames, as illustrated in Fig. \ref{fig:framework}(b). AMSA requires precise cross attention maps to achieve an excellent generation of different background and consistent characters across images. Acknowledging the constrained semantic representation of the original text encoder in Stable Diffusion, we introduce a Mask Perceptual Loss to improve the semantic representation of each character.

\paragraph{Auto-mask Space-Attention.}
Our aim is to ensure consistent character portrayal across inter-frame generation while integrating lively backgrounds. To achieve this, ID-Synchronizer extends the original self-attention module into a spatial self-attention module. Specifically, we rearrange the latent noise $z_{t}^{i,n}$ of each frame in the $i$-th layer of the diffusion model by formulating it as following:
\begin{equation}
\label{eq:zconcat}
    z_{t}^{i} = [z_{t}^{i,1} \oplus z_{t}^{i,2} \oplus...\oplus z_{t}^{i,N}]
\end{equation}
where $z_{t}^{i} \in \mathbb{R}^{(B \times N) \times H \times W \times C}$. Given that the self-attention mechanism in the diffusion model primarily handles visual information, we implement an auto-mask mechanism to incorporate attention masks of the main character region into spatial attention. This ensures that during the AMSA process, attention is masked, enabling each image to concentrate exclusively on the main character region of other frames within the batch.

In our task, the cross-attention maps are obtained to capture the areas of multiple  characters in the latent image. Considering maintaining the text alignment in the story generation task, we do not make any changes to the cross-attention modules in the diffusion model. During the training process, each self-attention layer receives cross-attention maps from all preceding layers. We capture the cross-attention map of each frame in a series sample by calculating between the text embedding of $P_{n}$ obtained in Eq. \ref{eq:Tn} and each noise image latent $z_{t}$ of $i$-th UNet layer following Eq. \ref{eq:attention_cal}:

\begin{equation}
\begin{aligned}
\label{eq:attention_cal}
    q_t^i &= W_{q}^{i} \cdot z_{t}^{i}, \quad
    k_n^i = W_{k}^{i} \cdot \mathcal{E}(P_{n}) \\
    m_{P_{n},t}^{i} &= \sum\limits_{i=1}^{i} \text{Softmax}(\frac{q_t^i \cdot k_n^i}{\sqrt{d_{k}}}),
    \quad n = 1,\ldots,N
\end{aligned}
\end{equation}

\noindent where $n$ denotes $n$-th frame mentioned in Eq. \ref{eq:textprompt}, $W_{q}^{i}$,$W_{k}^{i}$ are projection metrics in the cross attention module of the $i$-th layer, $\mathcal{E}$ represents the text encoder that encodes $P$ into text embeddings. Thus, the masks across the inter-frame collection are defined as follows:
\begin{equation}
\label{eq:mask_all}
    M_{P,t}^{i} = [m_{P_{1},t}^{i}\oplus m_{P_{2},t}^{i}\oplus...\oplus m_{P_{N},t}^{i}],
\end{equation}
where $n$ denotes each frame in a series of training samples, $i$ refers to the $i$-th layer of UNet. 

With the formulated latent noise $z_{t}^{i}$ in Eq. \ref{eq:zconcat} and the attention masks $M_{P,t}^{i}$ obtained by Eq. \ref{eq:mask_all}, the hidden states of $i$-th layer of the diffusion model are finally calculated as follows:
\begin{equation}
\begin{aligned}
    Q^{i} &= W_{q}^{i} z_{t}^{i}, K^{i} = W_{k}^{i} z_{t}^{i}, V^{i} = W_{v}^{i} z_{t}^{i} \\
    {z'}_{t}^{i} &= 
    Softmax( Q^{i} \cdot K^{i} / \sqrt{d_{k}} + \log {M_{P,t}^{i}}) \cdot V^{i}
\end{aligned}
\end{equation}
where $W_{q}^{i}$,$W_{k}^{i}$,$W_{v}^{i}$ are projection matrices,
${z'}_{t}^{i}$ is the new hidden states of $i$-th layer of UNet after AMSA. 

\paragraph{Mask Perceptual Loss} 
\begin{figure}[!t]
    \centering
    \includegraphics[width=0.6\linewidth]{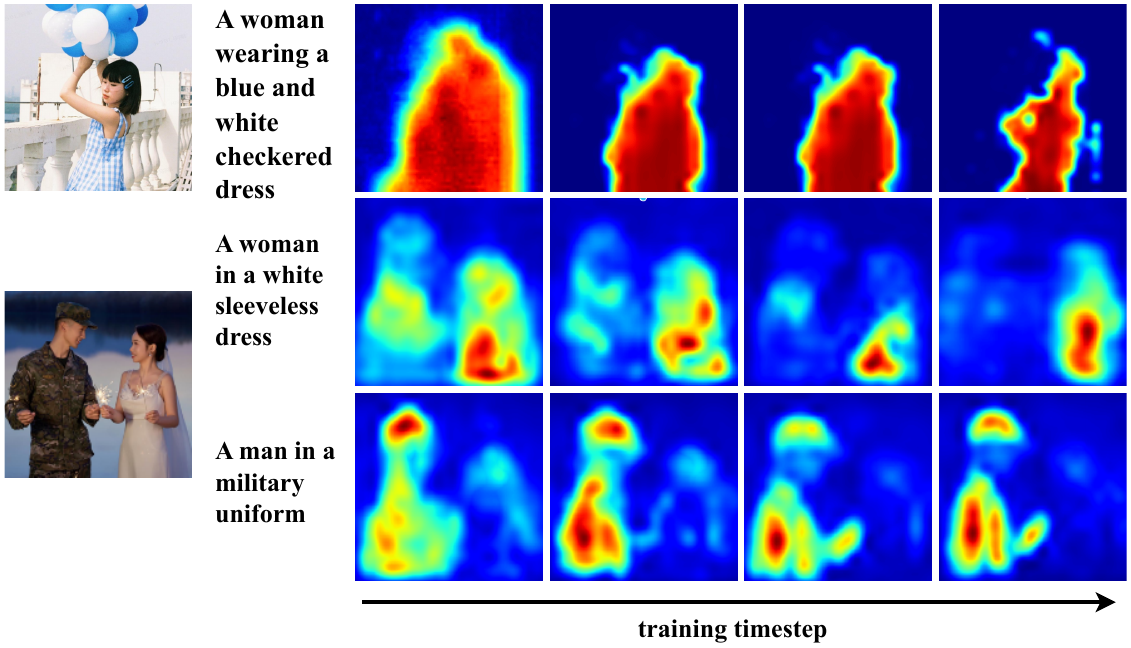}
    \caption{Cross attention map of each character during training. As the number of training steps increases, character attention maps gradually converge to accuracy within the constraints of mask perceptual loss.}
    \label{fig:maskattentionmap}
\end{figure}

AMSA's effectiveness relies on accurate cross-attention maps for high-quality, diverse background generation while maintaining character consistency. To enhance character semantic representation, we introduce a mask perceptual loss. We use a pre-trained segmentation model to obtain ground truth mask images for each character from training samples. Cross-attention maps are generated for each character and compared to the ground truth masks. We incorporate Dice loss\cite{sudre2017generalised} as an additional constraint to optimize cross-attention masks. Thus, the loss function is reconstructed as follows:
\begin{equation}
	\mathcal{L} = \mathcal{L}_{LDM} + \alpha{\sum\limits_{i=1}^{N}(1-\dfrac{2*\sum\limits_{i=1}^{M}p_{i}*g_{i}}{\sum\limits_{i=1}^{M}p_{i}^2+\sum\limits_{i=1}^{M}g_{i}^2})},
 \label{eq:LDM_LOSS}
\end{equation}

where $p_{i}$ refers the $i$-th pixel value of predict mask converted from $M_{\mathcal{T}_{tokens},t}^{k}$ and $g_{i}$ represents the $i th$ pixel value of ground truth mask images. $M$ is the total number of pixels. $N$ is the total characters in a training sample. $\mathcal{L}_{LDM}$ represents the original loss of latent diffusion models, $\alpha$ is the hyperparameter of the weight of mask loss. Fig. \ref{fig:maskattentionmap} illustrates the evolution of attention maps throughout the training process. Over the course of training, the cross-attention maps progressively become more accurate and increasingly resemble the ground truth masks.




\subsection{ID-Injector}
\begin{figure}[!t]
    \centering
    \includegraphics[width=1\linewidth]{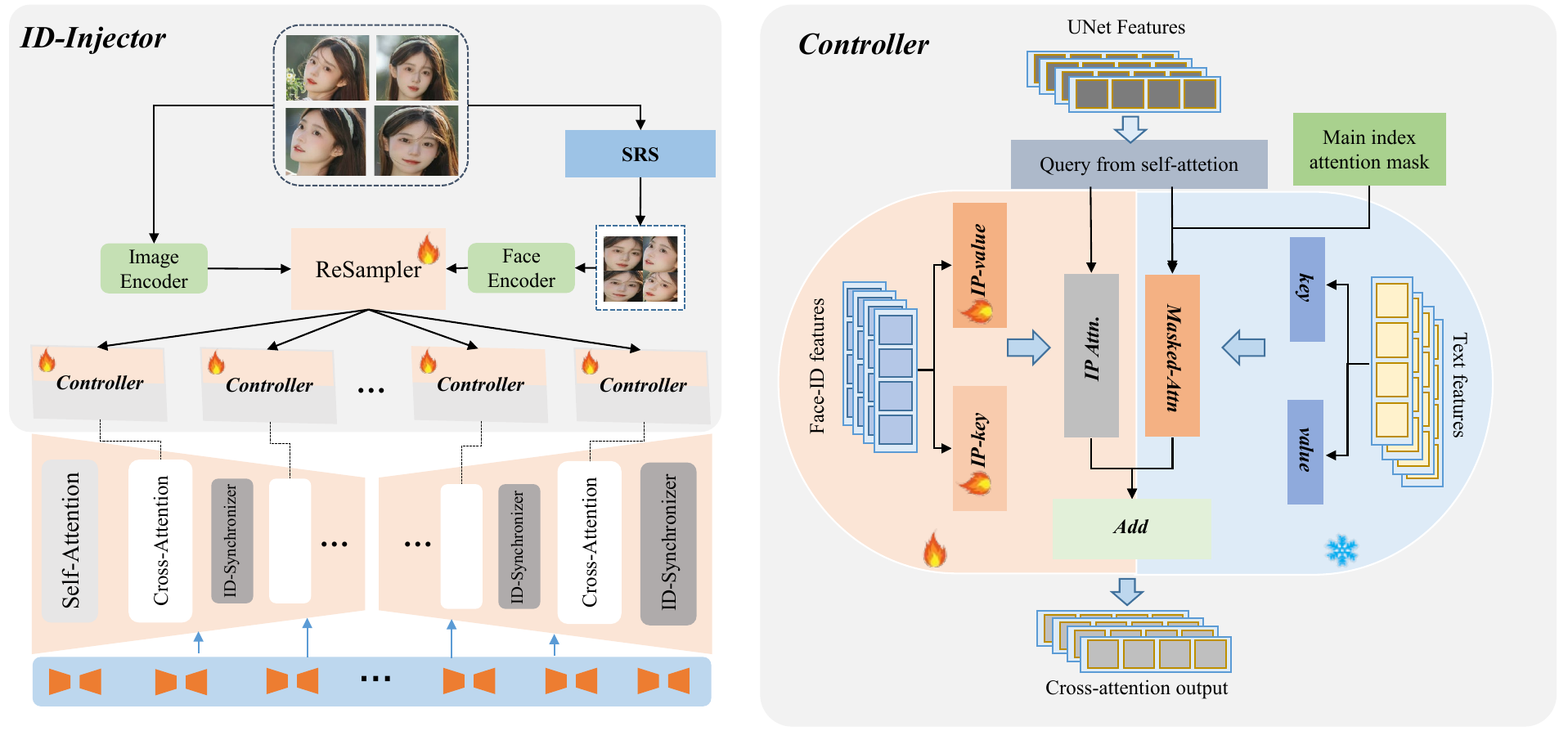}
    \caption{The structure of ID-Injector. The reference ID images are shuffled through \textbf{Shuffling Reference Strategy(SPS)}, enhancing the pose flexibility across frames. A Resampler and several inter-frame controllers are introduced to integrate reference ID images into the ID-Synchronizer.}
    \label{fig:ID-Injecter}
\end{figure}

Since the ID-Injector is trained alongside the ID-Synchronizer, it necessitates inter-frame feature injection. Given arbitrary numbers of ID images, Storynizor develops an optional inter-frame ID-Injector, which can receive additional face ID features for continuous story generation across frames. We adopt an ID encoder $\mathcal{E}_{f}$ to extract ID features from given face images $\mathcal{I}_{R}$ and a CLIP encoder $\mathcal{E}_{I}$ to extract image embeddings of this face. Then we develop a Resampler $\mathcal{P}_{r}$ to project the face images to the condition space of the latent diffusion model. Given a set of reference images $\mathcal{I}_{R}=\{\mathcal{I}_{n}, n=1,...,N$, the inter-frame face embedding finally into the diffusion model is defined as the following:
\begin{equation}
	c_{f} = \mathcal{P}_{r}(\mathcal{E}_{f}(\mathcal{I}_{R}),\mathcal{E}_{I}(\mathcal{I}_{R})),
 \label{eq:face_emb}
\end{equation}
where $c_{f} \in \mathbb{R}^{(B \times N) \times T \times h}$, $T \times h$ refers to the dimension of face condition embedding of each frame, $B \times N$ refers to the batch size and numbers of frames. Subsequently, another inter-frame cross-attention adaptive module is introduced into the latent diffusion model to support face images as prompts, illustrated in Fig. \ref{fig:ID-Injecter}(right).


\paragraph{Shuffling Reference Strategy (SRS).} 


Recent works \cite{li2023photomaker,xiao2023fastcomposer}  demonstrate various approaches to inject personalized features into diffusion models, such as original ID embedding, average ID embedding, stacked ID embedding and ID embedding with face keypoints. However, when used with ID-Synchronizer, with the integration of spatial attention modules in AMSA, the generated images are notably influenced by the initial image conditions, leading to consistent facial poses throughout the story generation process. Consequently, the generated facial poses in the images tend to align more closely with the input images.

We develop a new Shuffling Reference Strategy to our Storynizor. As illustrated in Fig. \ref{fig:ID-Injecter}(a), after packaging a set of reference images with the same ID, the SPS module is utilized to shuffle the set, resulting in a shuffled $\mathcal{I}_{R}^{'}$. Subsequently, injecting this shuffled set into the Resampler $\mathcal{P}_{r}$ yields a shuffled ID embedding.

Specifically, each training sample comprises $N$ images and $N$ associated prompts. We only consider single-character generation in training our ID-Injector. The training dataset contains:
\begin{equation}
\begin{aligned}
	\mathcal{I}_{R} = \{\mathcal{I}_{1},\mathcal{I}_{2},...,\mathcal{I}_{N}\} \\
\end{aligned}
 \label{eq:dataset}
\end{equation}
This bucket $\mathcal{I}_{R}$ can serve as a unified face condition space. During the training process, we shuffle the bucket $\mathcal{I}_{R}$ with the following:
\begin{equation}
	\mathcal{I}_{R}^{'} = \{\mathcal{I}_{s_1},\mathcal{I}_{s_2},...,
 \mathcal{I}_{s_N}\}
 \label{eq:shuffle}
\end{equation}
\noindent where ${s_n}$ indicates a shuffled index of the reference images.
Thus, Eq. \ref{eq:face_emb} can be written as the following to apply SPS into inter-frame ID-Injector:
\begin{equation}
	c_{f} = \mathcal{P}_{r}(\mathcal{E}_{f}(\mathcal{I}_{R}^{'}),\mathcal{E}_{I}(\mathcal{I}_{R}^{'})),
 \label{eq:face_emb_new}
\end{equation}
The feature set $c_{f}$ comprises a collection of individual ID features for each frame. Through the use of SPS, we can guarantee that every ID feature within $c_{f}$ is paired with another latent noise within the diffusion model. 

To inject $c_{f}$ into the ID-Synchronizer, we leverage the intrinsic cross-attention mechanism within the diffusion model, expanding it into an inter-frame generation as follows:
\begin{equation}
\begin{aligned}
    Q^{i} = W_{q}^{i} z_{t}^{i}, K^{i} = W_{k}^{i} c_{f}, V^{i} = W_{v}^{i} c_{f} \\
    {z'}_{t}^{i} = Softmax( Q^{i} \cdot K^{i} / \sqrt{d_{k}}) \cdot V^{i}
\end{aligned}
\end{equation}
where $W_{q}^{i}$,$W_{k}^{i}$,$W_{v}^{i}$ are projection matrices, ${z'}_{t}^{i}$ is the new hidden state of $i$-th layer of UNet after the inter-frame cross attention mechanism. 

In contrast to other methods, SPS allows each image to condition on a reference image with the same ID but different from itself. This unified representation significantly enhances the robustness of the facial pose in the generated images, particularly in inter-frame generation. 


\section{StoryDB Dataset Construction }

\begin{figure*}[!t]
    \centering
    \includegraphics[width=1\linewidth]{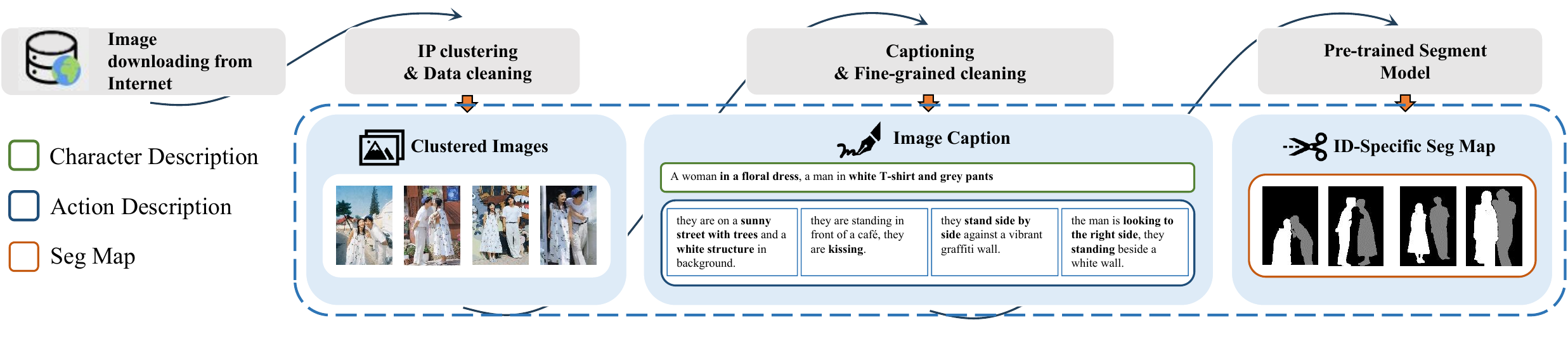}
    \caption{StoryDB Visualization and Data processing pipeline.}
    \vspace{-5mm}
    \label{fig:dataset}
\end{figure*}
Storynizor aims to generate consistent character images across diverse backgrounds. However, existing open-source datasets lack either rich background variety or fixed character attributes. To address this, we introduce \textbf{StoryDB}, a character-centric image-text pair dataset comprising 10,000 groups, each featuring the same character in consistent attire across different scenes, totaling 100,000 images. Each group contains 5-12 images with corresponding prompts, indexed shared prompt elements, and character mask images. StoryDB not only supports Storynizor's training but also serves as a resource for future research in story generation and IP-consistent content creation.

\textbf{Image downloading.} Initially, we collect images from the internet and open-source datasets to create a comprehensive character dataset comprising real humans, cartoon characters, and animals. We then calculate the aesthetic score of each image to aid in filtering the dataset during the download process.

\textbf{IP clustering.} We cluster the identical IPs to  generate several smaller datasets. Subsequently, We segment the images using category-specific keywords, and then calculate text-image score and image-image score using CLIP. Non-compliant samples are then filtered out based on these scores.

\textbf{Fine-grained filter and captioning.} We use GPT-4v to align and caption images within each category. Images are collectively input to GPT-4v for character alignment. Aligned images are captioned; non-compliant ones are rejected. GPT-4v labels image sets with the same character description. Finally, we manually correct non-compliant images in grouped sets to meet training dataset requirements.

\textbf{Tokenized and segmentation.} After getting the image text pairs, we extract the same description in the group prompts. This step holds significant importance as the identical description in the group prompts is crucial for generating the cross-attention map referenced in Equation \ref{eq:attention_cal}. Subsequently, We generate character mask images based on these descriptions using a pre-trained segmentation model called Segment Anything. These mask images serve as ground truth images to revise cross-attention maps during training.

\section{Experiments} \label{sec:experiments}

\subsection{Implementation Details}
We utilize the original checkpoint of Stable Diffusion Model-1.5 as the backbone for both ID-Synchronizer and ID-Injector. Training is conducted on 8 NVIDIA A100 GPUs, with 5\% probability of dropping out text and face conditions. Inference uses DDIM \cite{song2020denoising} with 30 steps and a guidance scale of 7.0 on an NVIDIA A30 GPU, with the resolution of 768 $\times$ 768.

\textbf{ID-Synchronizer.} We train the ID-Synchronizer with its UNet parameters frozen, using the StoryDB Dataset. We train 50,000 iterations with a batch size of 4 and learning rate of $5 \times 10^{-5}$ at a resolution of 512 $\times$ 512. The ID-Synchronizer is further fine-tuned at a resolution of 768 $\times$ 768 for high-fidelity generation, with a batch size of 1 for 50,000 iterations.

\textbf{ID-Injector.} We use a total of 80 million text-image pairs, comprising 50M from LAION-Face\cite{zheng2022general} and 30M from the internet. We train 2 epochs with a learning rate of $1 \times 10^{-4}$ and a batch size of 128 with the resolution of 512 $\times$ 512. In the second stage, we incorporate the pre-trained ID-Injector into Storynizor. We train the ID-Injector with ID-Synchronizer frozen, using the StoryDB for 5 epochs with a learning rate of $1 \times 10^{-4}$ and a batch size of 4, at the resolution of 768 $\times$ 768.



\subsection{Evaluation Dataset and Metrics}
We use GPT-4v to generate 100 character prompts and 100 story prompts, combining them randomly into 10k test groups. Each group contains 4-story prompts and 1 character prompt. We adopt CLIP-T for text-image alignment. CLIP-I and DINO-v2 \cite{oquab2023dinov2} are utilized to evaluate the similarity across inter-frame generated images. For ID-based generation, we randomly select 100 faces from FFHQ \cite{karras2019style} and use Arcface \cite{deng2019arcface} distance to evaluate the face similarity of the given image and the generated images (Face Sim(R)) and the face similarity among inter-frame generated images (Face Sim). 

\subsection{Quantitative Evaluation}
\begin{table}[!t]
\centering
\setlength\tabcolsep{1pt}
\scalebox{1}{
\begin{tabular}{cc|cccccc}
\toprule
Methods & Models & Clip-T$\uparrow$ & Clip-I$\uparrow$ & Dino-I$\uparrow$& Face Sim$\uparrow$ &Face Sim (R)$\uparrow$ &  \\
\midrule
\multirow{4}{*}{prompt-only} &                
                                
                                Storygen  &25.21           &67.45    &  67.42        &   10.82        &   -               \\
                                &Consistory      &  29.01        &  \underline{76.24}       &   \underline{79.22}       &  \underline{30.84}    &  -        \\
                                
                                 &Storydiffusion       & \underline{30.01}       & 72.56    &  70.34        & 23.44   &  -          \\
                                &\textbf{Storynizor}    & \textbf{33.28} & \textbf{83.33} & \textbf{86.62} & \textbf{41.55}& - &  \\
                                \midrule

\multirow{4}{*}{prompt-ID}

                              & IP-Adapter  &  28.26   &   66.43    &     65.83      &   26.57   &   20.57          \\
                               & PhotoMaker &  \textbf{32.46}   &   66.23    &    67.38       &   27.34   &    24.34             \\
                               & InstantID  &  25.44   &   \underline{79.46}    &    \underline{81.66}       &  \textbf{68.36}   &    \textbf{69.00}          \\
                               
                               & \textbf{Storynizor}  &  \underline{32.42}   &   \textbf{80.86}    &     \textbf{82.26}      &    \underline{39.64}  &           \underline{36.46}   \\

\bottomrule

\end{tabular}}
\caption{\label{tab:quantity}Quantitative results (\%) of Storynizor with other methods. Evaluations are conducted for both prompt-only and prompt-ID consistent story generation. The best and second-best results are highlighted in bold and underline, respectively. 
}
\end{table}

Quantitative results are presented in Tab. \ref{tab:quantity}. For prompt-only generation, our Storynizor achieves optimal performance in both text-image consistency and inter-image coherence. In prompt-ID guided generation, InstantID attains high scores in facial similarity. However, its semantic capability is compromised due to generating images overly similar to the reference, resulting in a lack of diversity, as evidenced by low CLIP-T scores. While PhotoMaker achieves comparable text similarity scores to our Storynizor, it significantly underperforms in story continuity and facial consistency. Overall, Storynizor demonstrates the highest comprehensive score, validating its superior story generation capabilities.


\subsection{Qualitative Evaluation}
\begin{figure*}[!t]
    \centering
    \includegraphics[width=1\linewidth]{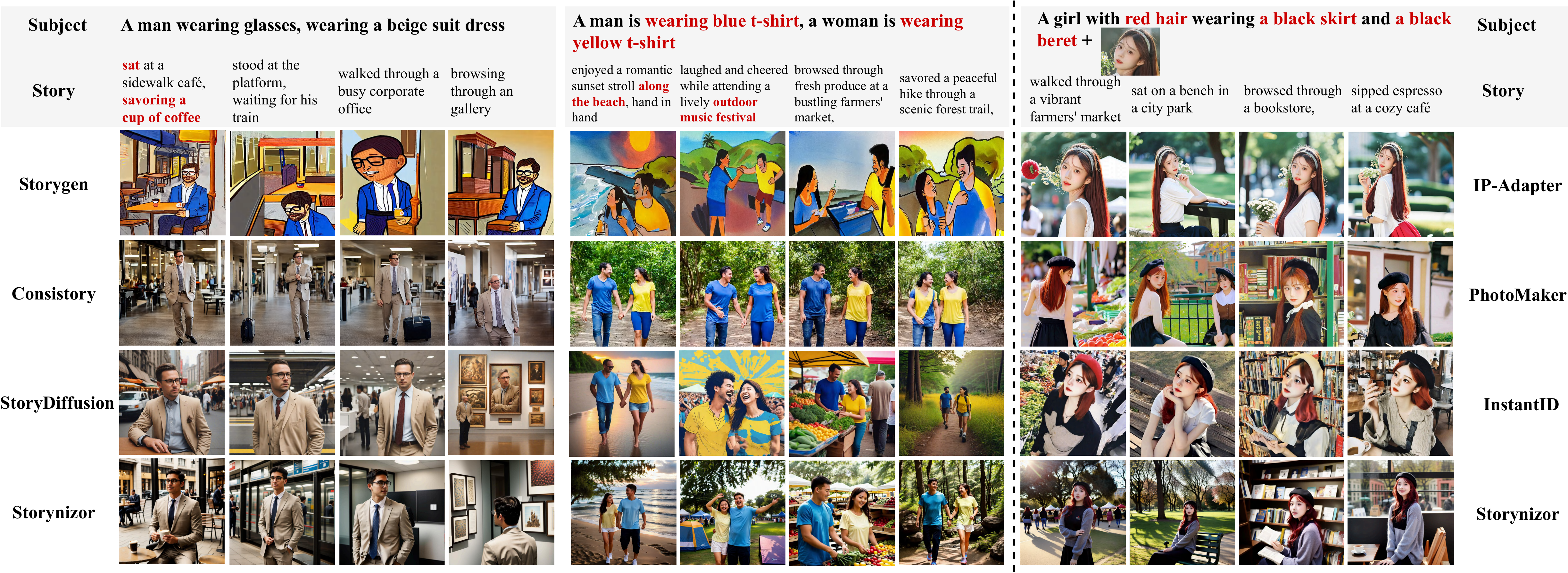}
    \caption{Qualitative comparison of Storynizor and other consistent story generation methods. We observe Storynizor outperforms other methods when generating consistent characters with vivid backgrounds and flexible poses in prompt-only story generation. Additionally, it achieves high-fidelity ID preservation in prompt-ID story generation.}
    \label{fig:visulization}
\end{figure*}

Fig. \ref{fig:visulization} presents qualitative comparisons of the results. Storynizor achieves superior consistency in details while simultaneously maintaining greater diversity compared to other methods. 
As shown in the multi-character generation example, the images generated by Storygen exhibit confusion in distinguishing between male and female attire, and lack text-image alignment. Consistory tends to produce similar character layouts across images while failing to clearly express character-specific semantic features. Storydiffusion similarly struggles with semantic ambiguity and demonstrates low consistency in preserving clothing details across images. In contrast, Storynizor achieves superior character consistency and background diversity in the generated images while ensuring semantic alignment. 
For prompt-ID guided generation, InstantID produces faces highly similar to the reference but lacks pose diversity and semantic fidelity. Similarly, IP-Adapter suffer significant semantic loss. While PhotoMaker generates characters from various angles, it falls short of Storynizor in narrative coherence.
Storynizor overcomes previous methods' limitations, generating coherent, diverse narratives while preserving reference ID consistency.


\subsection{Human Evaluation}
We conducted a user study with 25 experts to evaluate Storynizor against previous methods. Each expert evaluated the samples used for quantitative comparison. As shown in Table \ref{tab:user_study}, the results indicate a preference for Storynizor over other methods in both text alignments and consistent story generation.

\begin{table}[t]

   \centering
    \setlength{\tabcolsep}{0.8mm}{
    \begin{tabular}{cccccc}
     
     \toprule
      \multicolumn{1}{c}{Models} &
      \multicolumn{2}{c}{Text Alignments} & & \multicolumn{2}{c}{Consistent Generation}  \\
 \cmidrule{2-3} \cmidrule{5-6}
      (Storynizor vs. *) & Win(\%)  & Lose(\%) & & Win(\%) & Lose(\%)  \\
    \midrule
      IP-Adapter  & 91.2 & 8.8& & 69.8 & 30.2 \\
       InstantID & 100 & 0.0& & 100& 0.0 \\
       Photomaker & 74.8 & 25.2& & 79.3& 20.7 \\
       Storygen & 97.8 & 2.2& & 99.2& 0.8 \\
       Consistory & 69.2 & 30.8& & 61.7& 38.3 \\
      Storydiffusion & 65.8 & 34.2& & 63.2& 36.8 \\
      
    \bottomrule
     \end{tabular}}
     
     \caption{\label{tab:user_study}Human evaluation on Storynizor and other existing consistent story generation methods.}
 \end{table}

\subsection{Ablation Studies}

\textbf{Influence of AMSA and MPL of ID-Synchronizer.}
We conduct an ablation study of the following components: (1) Auto-Mask Self-Attention module (AMSA) and (2) Mask Perceptual Loss (MPL). Quantitative results are provided in Tab. \ref{tab:abl_mask}. As evidenced by the table, both the incorporation of AMSA and MPL results in notable improvements across all metrics for our model. The results are shown in Fig. With the utilization of AMSA, we observed a significant enhancement in character consistency of the generated results. Furthermore, with the incorporation of the mask loss, we observed a marked improvement in the consistency of fine details in our model's generated results, particularly evident in the clothing colors and accessories of the figures depicted in the images.

\textbf{Benefits of using SRS to shuffle the input IDs.}
Our ID-Injector incorporates personality features from given face images into cross-frame story generation. We conducted an ablation study to determine the optimal injection mode. Tab. \ref{tab:abl} shows that our proposed shuffling reference strategy (SRS) outperforms stacked ID embedding in both facial similarity and textual alignment, which corroborates the superiority of SRS.
\begin{table}[!t]
    \centering

    \scalebox{1}{
    \begin{tabular}{cc|cccccc}
    \toprule
    \makebox[0.1\textwidth][c]{AMSA} & \makebox[0.1\textwidth][c]{MPL} 
                                    & \makebox[0.1\textwidth][c]{CLIP-T$\uparrow$} & \makebox[0.1\textwidth][c]{CLIP-I$\uparrow$} & 
                                     \makebox[0.1\textwidth][c]{DINO-I$\uparrow$} &
                                     \makebox[0.1\textwidth][c]{Face Sim$\uparrow$} 
                                     \\

    \midrule
    \ding{55} & \ding{55}  & 30.46 & 78.83 &81.29 & 29.90  \\
     \ding{51} & \ding{55} & 32.51 & 81.66 & 83.74& 31.90   \\
     \ding{51} & \ding{51} &  \textbf{32.59}  &  \textbf{83.28}  & \textbf{85.58} & \textbf{36.55}& \\
    \bottomrule
    \end{tabular}
    }
    \caption{\label{tab:abl_mask} Quantitative ablation result (\%) of the components of our proposed ID-Sychronizer. AMSA stands for auto-mask self-attention, and MPL refers to mask perceptual loss. Each component is gradually added to evaluate its necessity and contribution to the overall performance. The experiments are conducted with the resolution of 512 $\times$ 512.}
\end{table}
\begin{figure}[!t]
    \centering
    \includegraphics[width=0.7\linewidth]{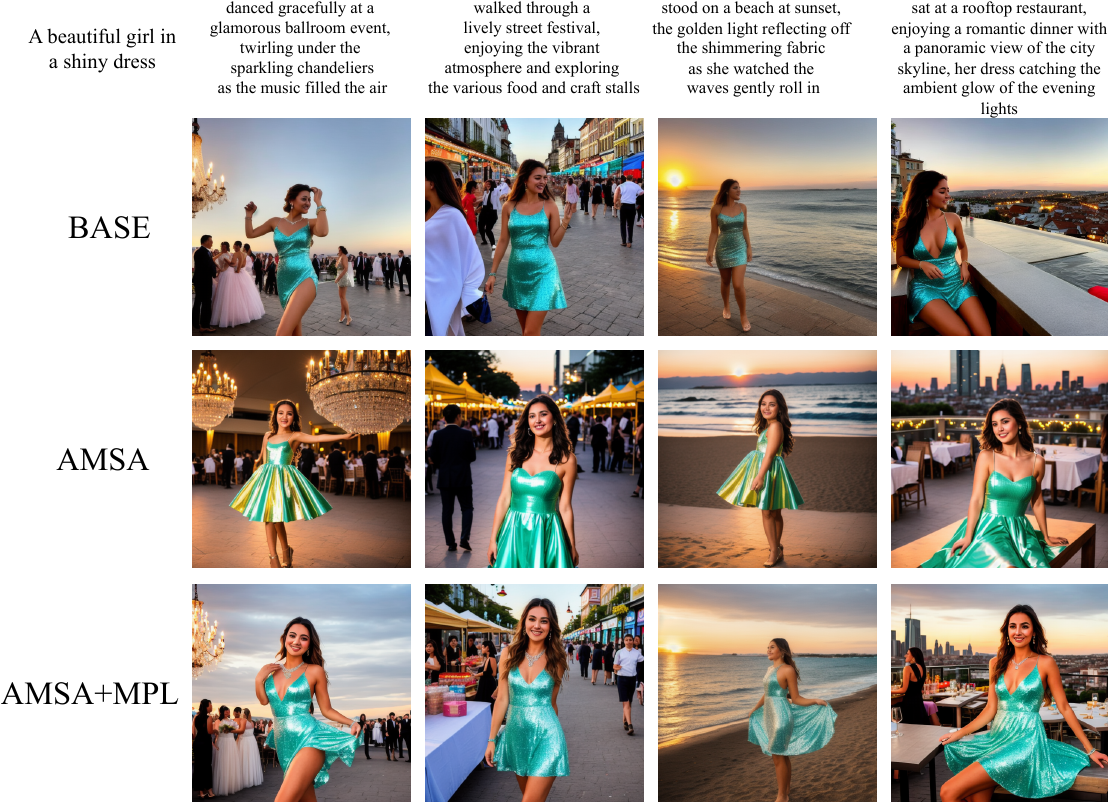}
    \caption{Qualitative ablation results of Storynizor with ASMA and MPL.
    }
    %
    \label{fig:comparison}

\end{figure}

\begin{table}[!t]
    \centering
    \small
    \scalebox{1}{
    \begin{tabular}{c|cccccccc}
    \toprule

    \makebox[0.05\textwidth][c]{Method} 
                                    & \makebox[0.1\textwidth][c]{CLIP-T$\uparrow$} & \makebox[0.1\textwidth][c]{CLIP-I$\uparrow$} & 
                                     \makebox[0.1\textwidth][c]{DINO-I$\uparrow$} &
                                     \makebox[0.1\textwidth][c]{Face Sim$\uparrow$} &
                                     \makebox[0.1\textwidth][c]{Face Sim(R)$\uparrow$} 
                                     \\

    \midrule
    Stacked-ID & 30.39 & 70.74&  72.71&19.26&17.72 \\
    SRS   & \textbf{32.59} & \textbf{71.65}&\textbf{75.63} & \textbf{36.48} & \textbf{32.57}   \\
    \bottomrule
    \end{tabular}}
    \caption{\label{tab:abl}Quantitative ablation result (\%) of different types of ID injections. Stacked-ID denotes that the reference ID image is identical to the latent image. SPS refers to our shuffle reference strategy.}
\end{table}

\section{Conclusion}


In conclusion, we present Storynizor, a model for generating cohesive story images with consistent characters, distinct foreground-background elements, and diverse poses. It combines ID-Synchronizer with AMSA for character consistency and vivid features, and the ID-Injector uses Shuffling Reference Strategy (SRS) for flexible face poses and consistent portrayal. Additionally, we introduce StoryDB, a 100,000-image dataset featuring diverse character sets in various settings, supporting Storynizor's training and future research.


\newpage
{
\small
\bibliographystyle{unsrt}
\bibliography{main}
}
\newpage
\appendix
\section{Appendix}
\label{appendix}

We have provided supplementary details regarding our Storynizor in this section. Our code will be released at https://anonymous.4open.science/r/Storynizor-0DC3. 

\subsection{Implementation details}
\label{implementaion}

\subsubsection{Inference setup}
\label{inference setup}
We utilize the original checkpoint of Stable Diffusion Model-1.5 as the backbone for both ID-Synchronizer and ID-Injector. Inference uses DDIM \cite{song2020denoising} with 30 steps and a guidance scale of 7.0 on an NVIDIA A30 GPU, with the resolution of 768 $\times$ 768.

\subsubsection{Training setup for ID-Synchronizer}
We train the ID-Synchronizer on 8 NVIDIA A100 GPUs. with its UNet parameters frozen, using the StoryDB Dataset. We train 50,000 iterations with a batch size of 4 and learning rate of $5 \times 10^{-5}$ at a resolution of 512 $\times$ 512. The ID-Synchronizer is further fine-tuned at a resolution of 768 $\times$ 768 for high-fidelity generation, with a batch size of 1 for 50,000 iterations.
\subsubsection{Training setup for ID-Injector}
We use a total of 80 million text-image pairs, comprising 50M from LAION-Face\cite{zheng2022general} and 30M from the internet. We train 2 epochs with a learning rate of $1 \times 10^{-4}$ and a batch size of 128 with the resolution of 512 $\times$ 512 on 8 NVIDIA A100 GPUs. In the second stage, we incorporate the pre-trained ID-Injector into Storynizor. We train the ID-Injector with ID-Synchronizer frozen, using the StoryDB for 5 epochs with a learning rate of $1 \times 10^{-4}$ and a batch size of 4, at the resolution of 768 $\times$ 768.

\subsubsection{Evaluation metrics}
\label{metrics}
We employ CLIP ViT-L/14\footnote{https://huggingface.co/openai/clip-vit-large-patch14} to evaluate the similarity between the generated images and the given text prompts (\textbf{CLIP-T}). Subsequently, we utilize the image encoder of the CLIP model to evaluate the correlation between the generated consistent images and the reference images (\textbf{CLIP-I}). Additionally, we employ the DINO score \cite{liu2023grounding} to evaluate image alignment, as DINO is better suited for subject representation (\textbf{DINO-I}). We use Arcface score to evaluate both the similarity between the generated faces and the reference face (Face Sim) and the similarity across the generated frames when evaluating the ID-Injector of Storynizor.




\subsubsection{Ablation}
\begin{figure}[ht]
    \centering
    \includegraphics[width=1\linewidth]{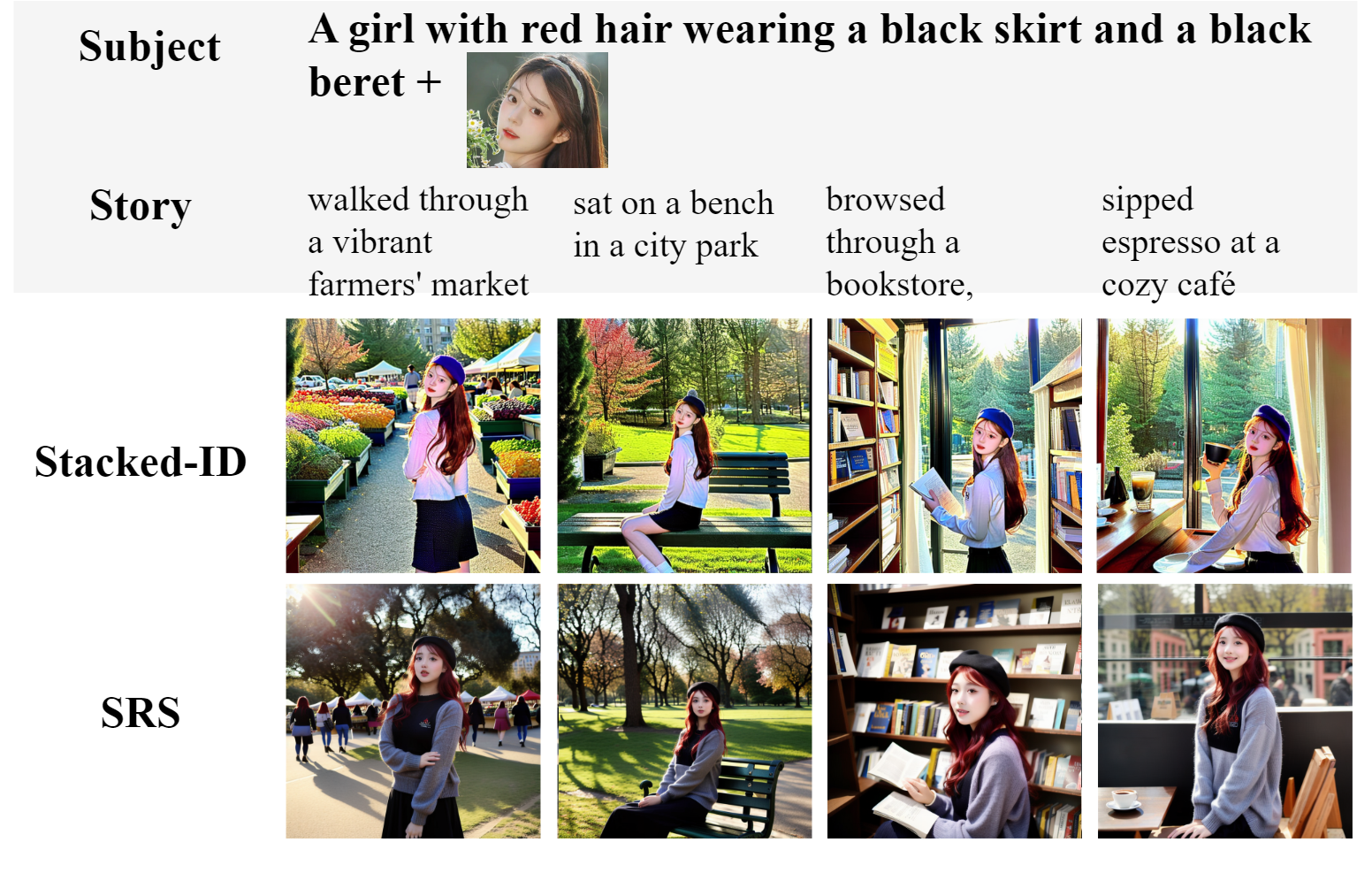}
    \caption{Qualitative ablation results of Storynizor with Stacked-ID embedding and shuffling reference strategy(SRS).}
    \label{fig:sps}
\end{figure}

We integrated ID features from given reference images into cross-frame story generation through our ID-Injector. An ablation study was carried out to identify the best injection mode. Each training sample contains four text-image-pairs. With Stacked-ID, faces from all training samples are stacked and injected into the Resampler during ID-Injector training, leading to stiff face postures, as illustrated in Fig. \ref{fig:sps}. In contrast, using our SPS resulted in more flexible face poses. Our quantitative results in the main paper also illustrate that our proposed shuffling reference strategy (SRS) outperforms stacked ID embedding in both facial similarity and textual alignment, affirming the superiority of SRS.

\begin{figure}[ht]
    \centering
    \includegraphics[width=1\linewidth]{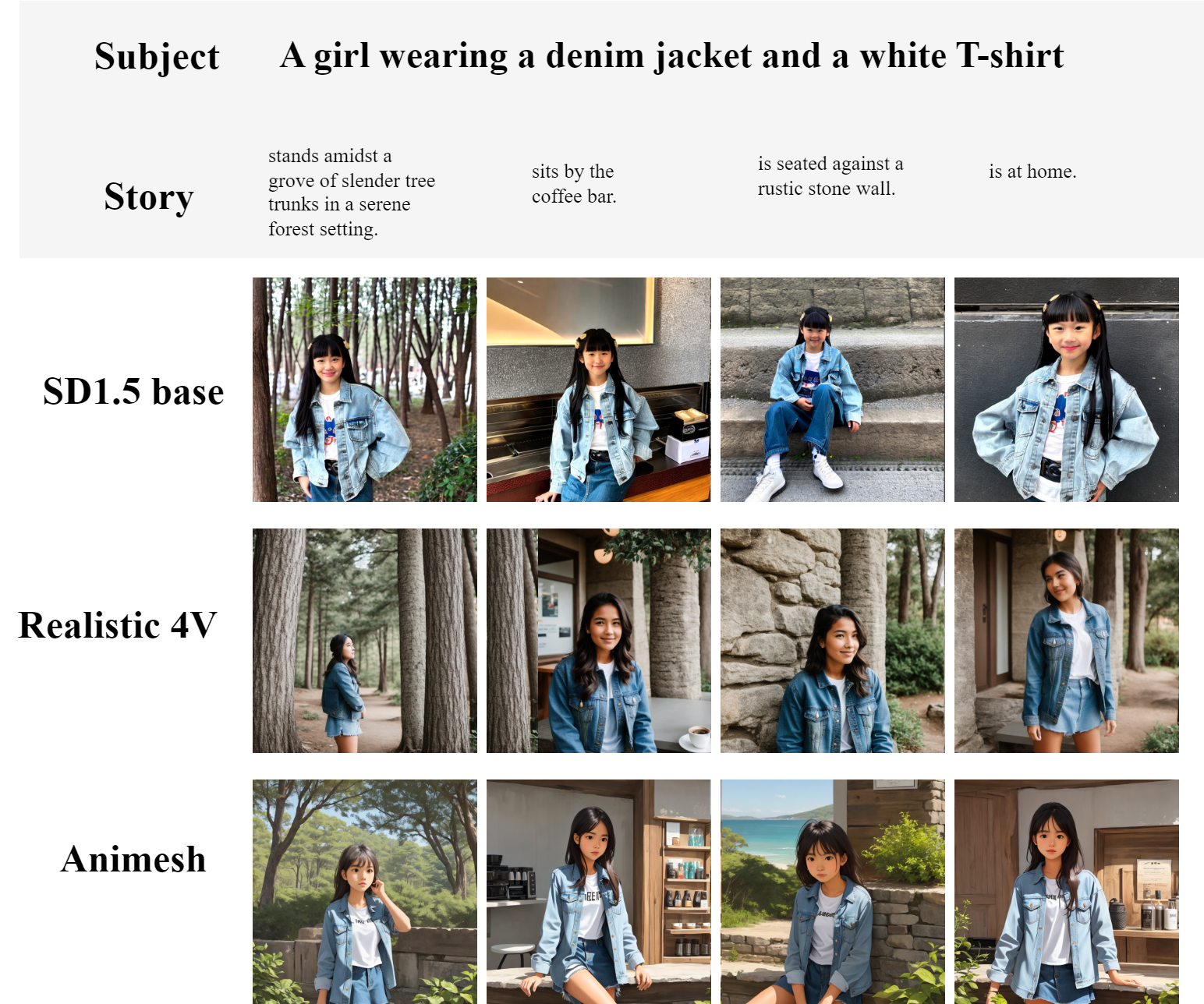}
    \caption{Qualitative additional results with different base models.}
    \label{fig:style-result}
\end{figure}

\subsubsection{More Visualization Results}

\begin{figure}[ht]
    \centering
    \includegraphics[width=1\linewidth]{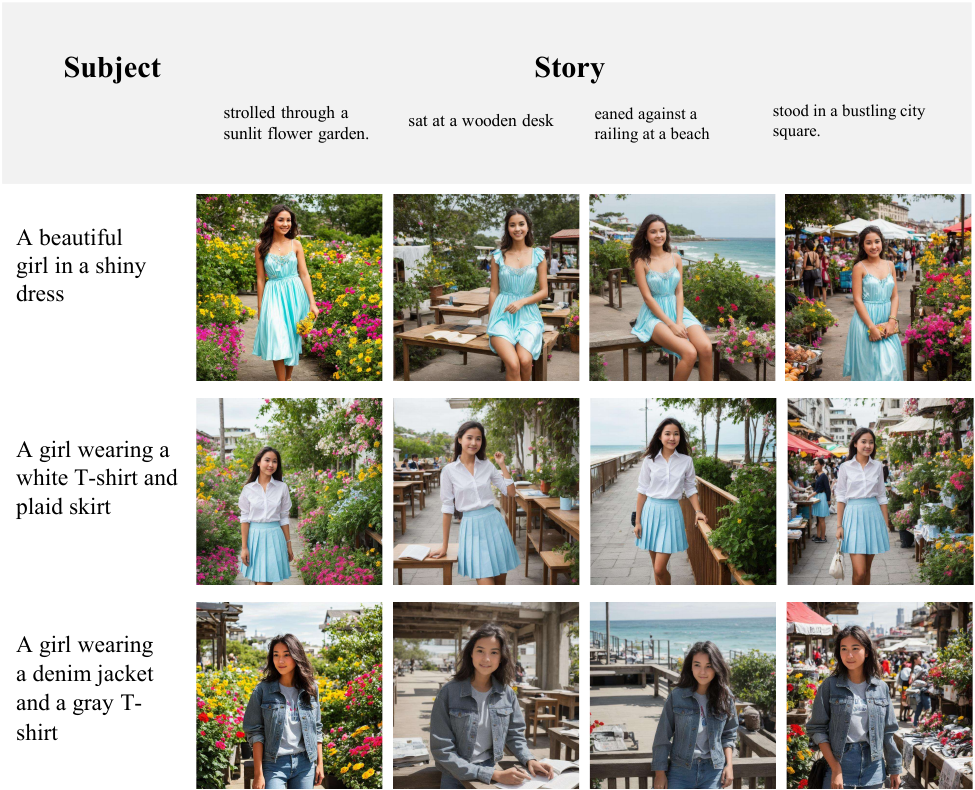}
    \caption{Qualitative additional results with single character.}
    \label{fig:single-result}
\end{figure}

\begin{figure}[ht]
    \centering
    \includegraphics[width=1\linewidth]{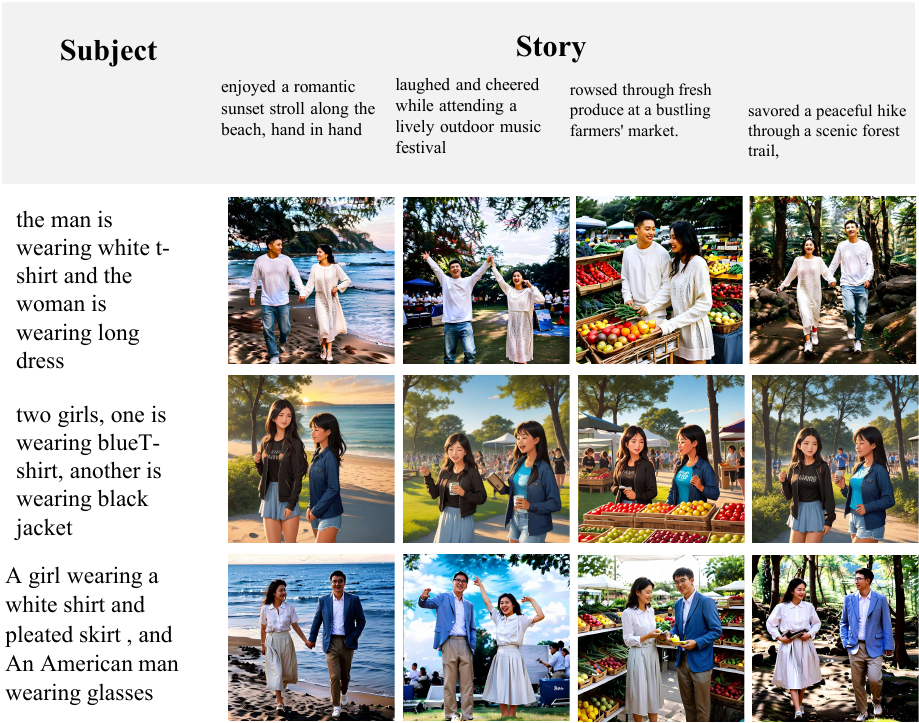}
    \caption{Qualitative additional results with multi characters.}
    \label{fig:double-result}
\end{figure}

\paragraph{Visualization for ID-Synchronizer}
As mentioned in the main paper, Storynizor can generate images with high consistent characters across frames, flexible postures and vivid backgrounds. Given a story and a prompt description of a character, Fig. \ref{fig:single-result} and Fig. \ref{fig:double-result} show the visualization results of single and multiple character generation of Storynizor, respectively. Furthermore, as is shown in Fig. \ref{fig:style-result}, our proposed Storynizor architecture is capable of integration with any diffusion model, facilitating the production of diverse stylized narratives.

\begin{figure}[htb]
    \centering
    \includegraphics[width=1\linewidth]{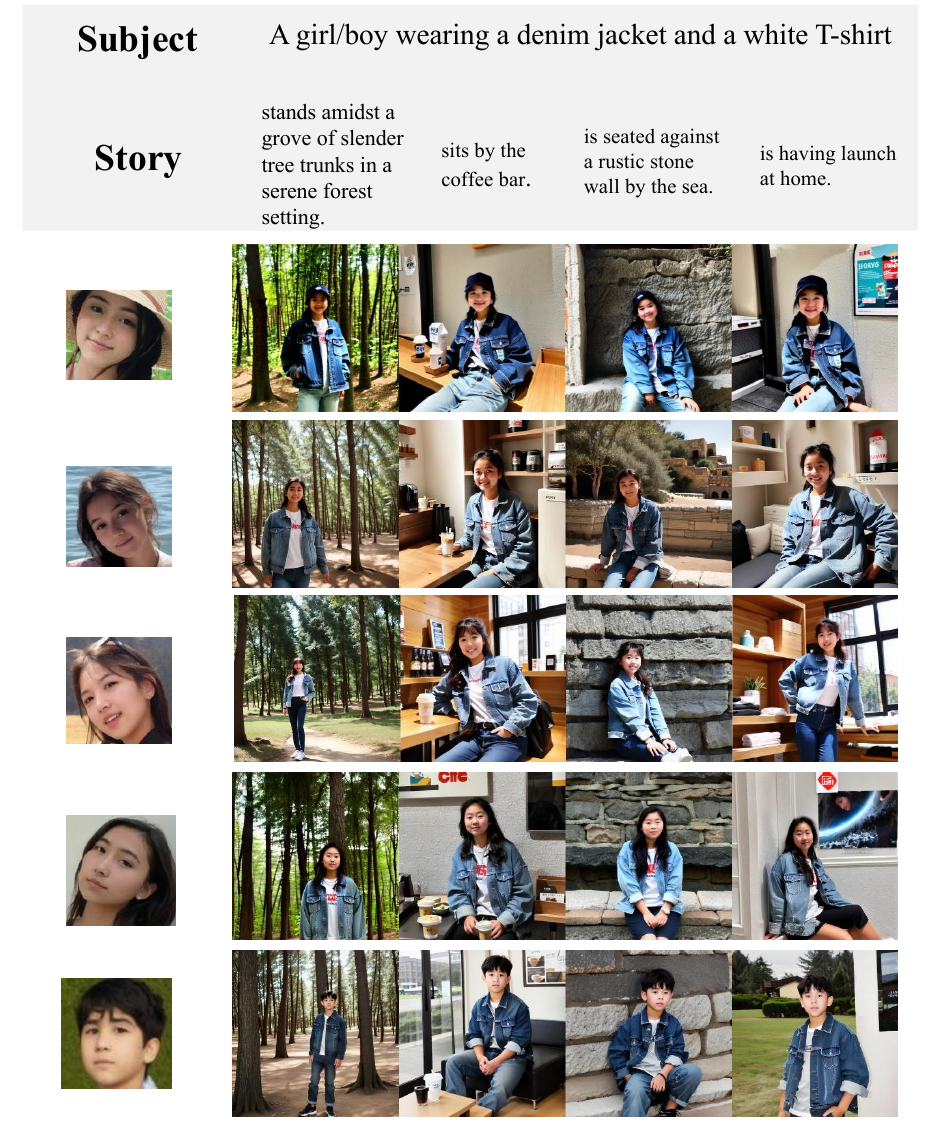}
    \caption{Qualitative additional results with ID conditions.}
    \label{fig:id-result}
\end{figure}

\paragraph{Visualization for ID-Injector}
Given a reference image, Storynizor is capable of generating images with high-fidelity ID-based consistent character generation as shown in Fig. \ref{fig:id-result}. High inter-frame story generation with a given character can be widely used in story telling, and continuous story generation to character development games.


\subsection{Limitations and discussion}
\label{limitaion}

While Storynizor is capable of generating stories with high inter-frame character consistency, effective foreground-background separation, and rich pose variation, several limitations warrant consideration. First, ID-Injector exclusively injects ID features into the ID-Synchronizer, supporting solely facial features and not other characteristics such as clothing. Clothing maintenance is uniquely handled within the ID-Synchronizer. To preserve character clothing based on a reference character, an Outfit-Injector can be included. We will leave the exploration as future work. Secondly, our method can only support multi-characters generation without reference images. When considering multi-character inputs, regional generation methods such as Character-Adapter\cite{ma2024characteradapterpromptguidedregioncontrol}, Mix-of-Show\cite{gu2024mix} and OMG\cite{kong2024omg} can be integrated with Storynizor. 

\subsection{Societal impacts}
\label{societal}
While our proposed method aims to deliver a versatile and powerful solution for creating stories with consistent character portrayal, effective foreground-background differentiation, and diverse pose variations, there are several limitations to consider. It can be widely used in story generation with high consistent characters. One important issue involves the potential misuse of the technology, which could lead to the creation of fabricated celebrity images, potentially causing public misinformation. It's worth noting that this concern is not specific to our approach, but is a shared consideration across all subject-driven image generation methods.

To address this, one possible solution involves implementing a safety checker similar to an NSFW filter, like the one found at https://huggingface.co/runwayml/stable-diffusion-v1-5, which functions as a classification module to assess whether generated images might be deemed offensive or harmful. This measure would serve to prevent the creation of controversial content and the misuse of celebrity imagery, thereby safeguarding against potential misuse of our method while upholding its intended purpose.

However, we acknowledge the ethical considerations arising from the ability to generate character images with high fidelity. The proliferation of this technology may lead to misuse of generated portraits, malicious image tampering, and an increase in the spread of false information. Therefore, we emphasize the importance of establishing and adhering to ethical guidelines and using this technology responsibly.



\end{document}